\documentclass[twocolumn]{article}
\usepackage[utf8]{inputenc}
\usepackage{graphicx}
\usepackage{tabularx}
\usepackage[a4paper,top=2cm,bottom=2cm,left=1.7cm,right=1.7cm]{geometry}
\usepackage[colorinlistoftodos]{todonotes}
\usepackage{multirow}
\usepackage{tikz}

\usepackage{ wasysym }
  
\title{Increasing Generality in Machine Learning through\\ Procedural Content Generation}
\author{
Sebastian Risi$^{1,2}$ and Julian Togelius$^{1,3}$\\
$^{1}$modl.ai, $^{2}$IT University of Copenhagen, $^{3}$New York University\\
\{sebastian, julian\}@modl.ai
}
\date{}

\begin{document}
 

\maketitle
Procedural Content Generation (PCG) refers to the practice, in videogames and other games, of generating content such as levels, quests, or characters algorithmically. Motivated by the need to make games replayable, as well as to reduce authoring burden, limit storage space requirements, and enable particular aesthetics, a large number of PCG methods have been devised by game developers. Additionally, researchers have explored adapting methods from machine learning, optimization, and constraint solving to PCG problems. Games have been widely used in AI research since the inception of the field, and in recent years have been used to develop and  benchmark new machine learning algorithms. Through this practice, it has become more apparent that these algorithms are susceptible to overfitting. Often, an algorithm will not learn a general policy, but instead a policy that will only work for a particular version of a particular task with particular initial parameters. In response, researchers have begun exploring randomization of problem parameters to counteract such overfitting and to allow trained policies to more easily transfer from one environment to another, such as from a simulated robot  to a robot in the real world. Here we review the large amount of existing work on PCG, which we believe has an important role to play in increasing the generality of machine learning methods. The main goal here is to present RL/AI with new tools from the PCG toolbox, and its secondary goal is to explain to game developers and researchers a way in which their work is relevant to AI research.

\section{Introduction}
For several decades, procedural content generation (PCG) has been a feature of many video games (Table.~\ref{tbl:overview}). PCG refers to the algorithmic creation of game content --- not the game engine, but things such as levels, quests, maps, characters or even rules --- either in runtime (as the game is being played) or at design time (as the game is made). There are several reasons why PCG is used in games: it can increase the replayability of a game as players are presented with a new experience every time they play, it can help to reduce production costs and disk storage space, and it enables new types of games built on the unique affordances of content generation.

Interestingly, developments in PCG and machine learning have started to influence each other in reciprocal ways. \emph{Procedural Content Generation
via Machine Learning} (PCGML) \cite{summerville2018procedural} refers to the use of machine learning to train models on existing game content, and then leverage these models to create novel content automatically. This can be done through simply sampling from the learned models, or through searching the artifact  space implied by the model so as to optimize some objective. In this paper, the term artifact refers to objects/levels/maps/etc. made by an algorithm. Interestingly, PCGML poses different and hard challenges compared to generating e.g.\ images, because the produced content needs to function.

At the same time as PCG researchers are starting to incorporate these advances into their systems, interest in the machine learning community is increasing in PCG-inspired methods to improve the  robustness of ML systems. One reason for this development is the growing evidence that, while ML methods perform well for tasks or in the environments they are trained on, they do not generalize well when that environment is changed or different from what is seen during training. Training neural networks with many free parameters and over long training times has lead to  state-of-the-art performance in many domains, but these solutions are typically overfitted to the particular training examples, achieving high accuracy on a training set but performing poorly on data not used for training \cite{hardt2016equality,lin2016generalization}. 
Especially in deep reinforcement learning (in which an agent has to learn in interaction with its environment), overfitting is rarely addressed \cite{justesen2018illuminating,zhang2018study,ruderman2018uncovering} but a significant problem. 
Take the very popular Arcade Learning Environment (ALE) as an example \cite{bellemare2013arcade}, a classic benchmark in RL based on an emulation of the Atari 2600 games console. Hundreds of games were made for that console, however they all have fixed sets of levels and very little in the way of randomization. Training an agent to play a game in ALE makes it liable to overfit not only to that particular game, but to its levels and sequence of events.

The basic idea in employing PCG to address the generality problem in ML systems is to artificially create more training data or training situations. This way ML-based systems can be biased towards learning general task properties instead of learning spurious elements found in the training examples. Methods include   simpler  approaches such as data augmentation that artificially increase the data used for training  \cite{krizhevsky2012imagenet,simard2003best} or methods that train agents in a large number of training environments that include randomized elements  \cite{tobin2017domain}. PCG methods have been extended to create complete maps for Capture the Flag \cite{jaderberg2019human} or maps for 2D video games \cite{torrado2019bootstrapping}, and in a recent impressive demonstration of the advantage of training in PCG-generated  environments, have  allowed a robot hand trained in a simulation  to manipulate a Rubik's cube in the real world \cite{openai2019solving}. 

In this article, we review the history of PCG and the recent trends in hybridizing PCG methods with machine learning techniques. We do this because we believe there are convergent research interests and complementary methods in the two communities. We want to supply machine learning researchers with a new toolbox to aid their generalization work, and games researchers and developers with new perspectives from machine learning. This Review also details promising future research direction and under-explored research avenues enabled by more advanced PCG techniques. The goal of this Review is to share with the larger machine learning research community work from the exciting field of PCG that is just beginning to capture the interest of ML researchers but could ultimately encourage the emergence of more general AI. We focus on examples where there's a notion of ``environment'' (typically through learning being centered on an agent in simulated physical world), but will throughout the text mention other learning settings where relevant for comparison.

We will discuss work coming out of both commercial game development, AI/ML research targeted at games, and AI/ML research targeted at other applications or the development of general intelligence. It is important to bear in mind that these approaches were developed for very different purposes, with those coming out of game development and research typically focused on providing entertainment. However, this should not in general make them less suitable for other purposes. Indeed, as one of the reasons games are entertaining is that they challenge our minds, one could argue that algorithms designed to increase entertainment in games may be useful for creating intelligence-relevant challenges~\cite{koster2005theory}.
\begin{figure*}
\centering
\includegraphics[width=1.0\textwidth]{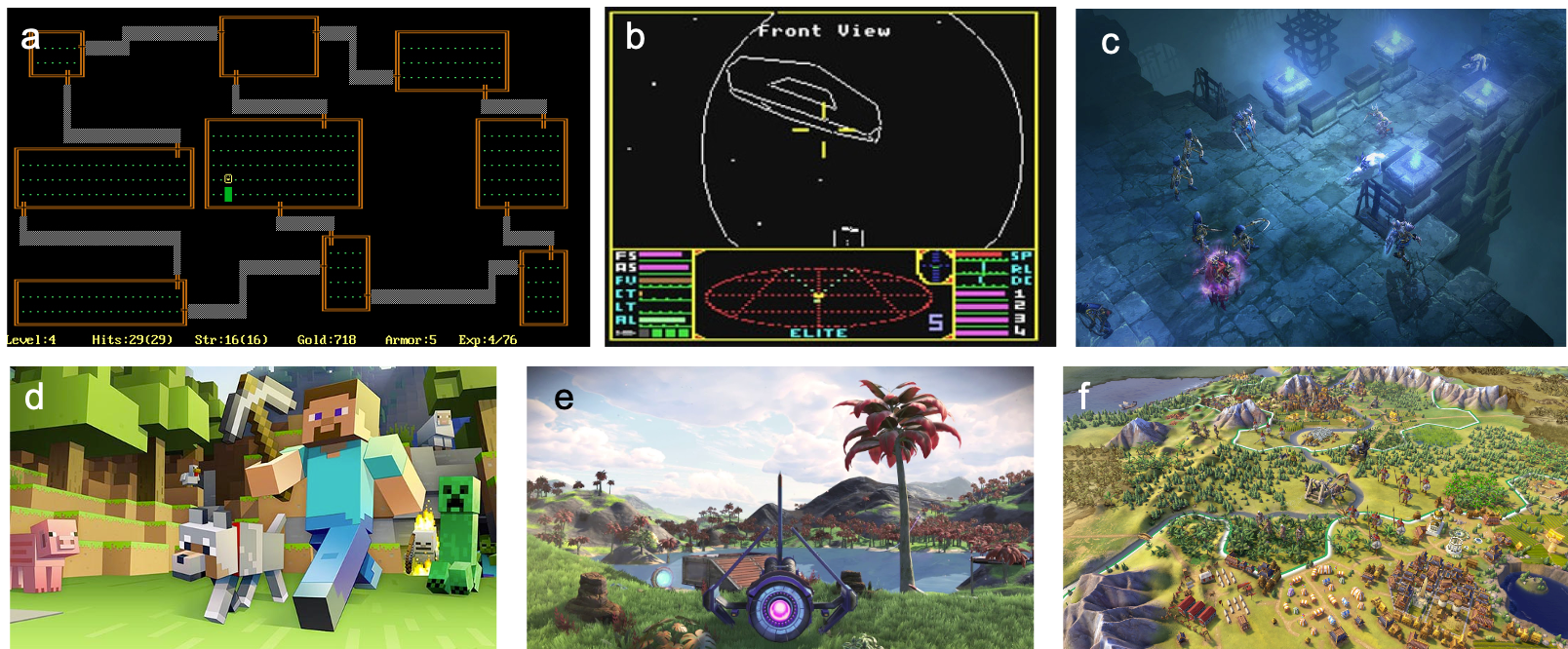}
\caption{Example commercial games that feature PCG as an important game component are Rogue (\textbf{a}), Elite (\textbf{b}), Diablo III (\textbf{c}), Minecraft (\textbf{d}), No Man's Sky (\textbf{e}), and Civilisation VI (\textbf{f}). }
\label{fig:pcg_commercial}
\end{figure*} 

We will discuss both work that we classify as PCG and certain examples of work that is adjacent to PCG, such as some forms of data augmentation in supervised learning. Our necessarily imperfect line of demarcation is that pure randomization/shuffling is not PCG; however, many PCG algorithms include randomness.


\begin{table*}[]
\label{tbl:overview}
\caption{A comparison of several methods for PCG and domain randomization described in this article. The first two columns indicate the representation of the content, which is either designed by hand or learned through machine learning. The last columns indicate how the content is generated given a representation. Constructive methods follow rules and do not do any resampling. Dwarf Fortress is an example of a generate-and-test (random search) method where the world is regenerated if it fails certain tests. Most PCGML approaches randomly sample a learned representation, whereas PCGML with constraints resample when constraints are not satisfied. In the search-based paradigm, a hand-coded representation is searched using an evolutionary algorithm. Latent Variable Evolution combines search-based PCG with a learned representation, e.g.\ in the form of a GAN. PCGRL uses a policy learned by reinforcement learning to search a hand-coded representation, whereas Generative Playing Networks instead uses reinforcement learning to test the artifacts and gradient descent to generate them. Activation Maximization relies on gradient descent to generate artifacts, based on a learned representation. The various forms of domain randomization use hand-coded representations and differ in whether they simply sample this space or perform some sort of search with a learned policy. POET and MCC are fundamentally search-based methods, which include a learning agent inside the evaluation loop. Progressive PCG uses a parameterizable constructive generator, coupled to a RL-based game-playing agent. }
\begin{tabular}{c|c|c|c|c|c|c|c|c|}
\cline{2-9}
 &
  \multicolumn{2}{c|}{Representation} & \multicolumn{6}{c|}{Generation Method} \\  \cline{2-9}
&  Learned & Hand- & Evol. & Learned &  Gradient- & Rand. & Sampl. & Rules  \\ 
&   & designed &
   & 
 & based & search &  &   \\ 
  \hline 
\multicolumn{1}{|l|}{Standard Constructive }              &    &  &   &   &   &   &    &  \\ 
\multicolumn{1}{|l|}{E.g.\ Rogue, Pitfall!,}            &  \fullmoon  & \newmoon & \fullmoon  & \fullmoon  & \fullmoon  &\fullmoon   &  \fullmoon  & \newmoon \\ 
\multicolumn{1}{|l|}{Civilization, Elite, Minecraft} &   &  &   &  &  &  &   &  \\ 
\hline
\multicolumn{1}{|l|}{Dwarf Fortress}  & \fullmoon & \newmoon   & \fullmoon & \fullmoon &\fullmoon  & \newmoon  &  \fullmoon &  \fullmoon  \\ \hline
\multicolumn{1}{|l|}{Standard PCGML \cite{summerville2018procedural,summerville2016mystical,summerville2016super}} & \newmoon  &\fullmoon  & \fullmoon  & \fullmoon & \fullmoon & \fullmoon & \newmoon  &  \fullmoon  \\ \hline
\multicolumn{1}{|l|}{PCGML with constrained } & \newmoon  &  \fullmoon & \fullmoon  & \fullmoon & \fullmoon & \newmoon  & \fullmoon &   \fullmoon \\ 
\multicolumn{1}{|l|}{sampling \cite{snodgrass2016controllable}} &  &  &   &  &  &  &  &   \\ \hline
\multicolumn{1}{|l|}{Standard search-based } &  \fullmoon & \newmoon   & \newmoon  & \fullmoon & \fullmoon & \fullmoon & \fullmoon  &   \fullmoon \\ 
\multicolumn{1}{|l|}{\cite{togelius2011searchbased,dahlskog2014multi,browne2010evolutionary,togelius2013controllable,togelius2007towards}}&  &   &  &  &  &  &   &   \\ \hline
\multicolumn{1}{|l|}{Latent Variable Evolution}  & \newmoon  &  \fullmoon  & \newmoon  &   \fullmoon &  \fullmoon &  \fullmoon &  \fullmoon  &  \fullmoon  \\ 
\multicolumn{1}{|l|}{ \cite{volz2018evolving,bontrager2018deepmasterprints}}  &  &   &  &  &  &  &   &   \\ 
\hline
\multicolumn{1}{|l|}{PCGRL \cite{khalifa2020pcgrl}}  &  \fullmoon  & \newmoon  & \fullmoon & \newmoon  & \fullmoon & \fullmoon &  \fullmoon &   \fullmoon \\ \hline
\multicolumn{1}{|l|}{Generative Playing}  & \newmoon  & \fullmoon & \fullmoon & \fullmoon & \newmoon  & \fullmoon & \fullmoon  & \fullmoon  \\ 
\multicolumn{1}{|l|}{Networks  \cite{bontrager2020fully}}  &  &  &  &  &  &  &   &   \\
\hline
\multicolumn{1}{|l|}{Activation Maximization \cite{nguyen2016synthesizing}}  & \newmoon  &  \fullmoon  &   \fullmoon &  \fullmoon & \newmoon    &  \fullmoon &  \fullmoon  &   \fullmoon \\ \hline

\multicolumn{1}{|l|}{Simple Data Augmentation }  & \newmoon   &  \fullmoon & \fullmoon & \fullmoon & \fullmoon & \fullmoon &  \fullmoon &  \newmoon  \\
\multicolumn{1}{|l|}{\cite{krizhevsky2012imagenet,simard2003best}
 }  &  &   &  &  &  &  &   &   \\
\hline
\multicolumn{1}{|l|}{Uniform Domain  }  & \fullmoon  & \newmoon   & \fullmoon & \fullmoon & \fullmoon & \fullmoon & \newmoon   &   \fullmoon \\ 
\multicolumn{1}{|l|}{Randomization \cite{tobin2017domain} }  &   &   &  &  &  &  &   &   \\
\hline
\multicolumn{1}{|l|}{Guided Domain  }  & \fullmoon & \newmoon  & \fullmoon & \newmoon  & \fullmoon & \fullmoon &  \fullmoon &  \fullmoon \\ 
\multicolumn{1}{|l|}{Randomization \cite{cubuk2018autoaugment} }  &  &   &  &  &  &  &   &   \\ 
\hline
\multicolumn{1}{|l|}{Automatic Domain}  & \fullmoon & \newmoon   & \fullmoon & \fullmoon & \fullmoon & \fullmoon &  \newmoon  &  \fullmoon \\ 
\multicolumn{1}{|l|}{Randomization \cite{openai2019solving}}  &  &   &  &  &  &  &   &   \\ \hline
\multicolumn{1}{|l|}{POET \cite{wang2019paired}, MCC \cite{brant2017minimal}}& \fullmoon & \newmoon  & \newmoon  & \fullmoon & \fullmoon & \fullmoon & \fullmoon  &  \fullmoon \\ 
\hline
\multicolumn{1}{|l|}{Progressive  PCG (PPCG) \cite{justesen2018illuminating} }& \fullmoon & \newmoon  &  \fullmoon & \fullmoon & \fullmoon & \fullmoon & \fullmoon & \newmoon  \\ 
\hline
\end{tabular}
\end{table*}

\section{Classic Procedural Content Generation}
\label{sec:classic_pcg}
While possibly the first video game to include PCG dates from
1978 (Beneath Apple Manor by Don Worth for the Apple
II), Rogue (1980) by Toy and Wichmann created an important design paradigm. In Rogue (Fig.~\ref{fig:pcg_commercial}a), the player explores a multi-level dungeon complex, battling enemies and collecting treasures. As the creators did not want to author the dungeons themselves (they wanted to play the game and be surprised), they needed to create a dungeon generation algorithm; every time you play a game of Rogue, a new set of dungeons are generated. Rogue came to inspire a genre of games called \emph{rogeuelikes}, which are characterized mainly by the use of runtime generation of content that is essential to gameplay. The highly successful {Diablo} series of games (Blizzard, 1997-2013) (Fig.~\ref{fig:pcg_commercial}c), as well as platformers such as \emph{Spelunky} (Mossmouth, 2008), are roguelikes. 

While the PCG in Rogue was motivated by a need for replayability and unpredictability, another key reason for PCG is wanting to create game worlds larger than can fit in memory or on storage media. A   paradigm-setting game here was \emph{Elite} (Brabensoft, 1984), a spacefaring adventure game featuring thousands of planets which seemingly miraculously fit in memory on a Commodore 64, with 64 kilobytes of memory (Fig.~\ref{fig:pcg_commercial}b). Every time a star system was visited, the game would recreate the whole starsystem with planets, space stations, and spacecraft, from a given random seed. This approach has later been used for games such as \emph{No Man's Sky} (Hello Games, 2015), which famously contains more planets than you can visit in a lifetime, all with their own ecologies (Fig.~\ref{fig:pcg_commercial}e).

The strategy games in the very popular \emph{Civilization} series also rely heavily on PCG, as a new world is created for the players to explore and contest every time a new game is created (Fig.~\ref{fig:pcg_commercial}f). Similarly, the open-world sandbox game \emph{Minecraft} (Mojang, 2010) creates a completely new world at the start of each game session (Fig.~\ref{fig:pcg_commercial}d). Other games use PCG in somewhat more peripheral roles, such as the sidequest generation (e.g.\ creating an infinite supply of fetch quests  through a guild system) in \emph{The Elder Scrolls V: Skyrim} (Bethesda, 2011) (along with some earlier games in the series) and the pervasive generation of terrain features and vegetation in a large number of open-world 3D games. PCG techniques are now so commonplace and reliable that it is more common than not to utilize them in many game genres.

Interestingly, PCG in video games is actually prefigured by certain pen-and-paper generators intended to be executed by humans with the help of dice or cards, including a dungeon generator for the classic \emph{Dungeons and Dragons} (TSR, 1976) role playing game~\cite{smith2015analog}. Some recent board games which include aspects of PCG are \emph{504} (2F Spiele, 2015) or \emph{Betrayal at House on the Hill} (Avalon Hill, 2004).

The types of PCG that can be found in most existing games are called \emph{constructive} PCG methods (Table~\ref{tbl:overview}). This means that the content generation algorithm runs in a fixed time, without iteration, and does not perform any search. 
For generating textures, heightmaps, and similar content, a commonly used family of algorithms are fractal noise algorithms such as Perlin noise~\cite{perlin1985image}. Vegetation, cave systems, and similar branching structures can be efficiently generated with graphically interpreted grammars such as L-systems~\cite{prusinkiewicz1986graphical}. Other constructive methods that are borrowed from other fields of computer science and adapted to the needs of PCG in games in include cellular automata~\cite{von1951general} and other approaches based on local computation. Other constructive methods are based on rather less principled and more game-specific methods. For example, \emph{Spelunky} combines a number of pre-authored level chunks according to patterns which are designed so as to ensure unbroken paths from entrance to exit. 


\section{``PCG'' in Machine Learning: Data Augmentation and  Domain Randomization}
\label{sec:augment}

While not necessarily called PCG in the machine learning community, the idea of \emph{data augmentation} is essentially a simple form of constructive PCG. The aim of data augmentation is to increase the diversity in the dataset, not by collecting more data but by adding modified versions  of the already existing data \cite{krizhevsky2012imagenet,simard2003best}. Data augmentation is very common in supervised learning tasks, for example, through cropping, padding, or adding noise to images in a dataset. It is common practice in machine learning and has resulted in significantly less overfitting and state-of-art results in a variety of domains \cite{krizhevsky2012imagenet,perez2017effectiveness,cui2015data}.  

A different form of data augmentation was introduced by Geirhos et al.~\cite{geirhos2018imagenet}, in which the authors showed that training the \emph{same} network architecture but with a stylized version of ImageNet images (e.g.\ a cat with the texture of an elephant) can significantly increase the model’s accuracy and robustness. In fact, the authors showed that a deep convolutional network trained on the standard ImageNet mainly focuses on textures in images instead of their shape; training on the stylized version of ImageNet increases their shape bias and with that, their accuracy and robustness. 

In the field of reinforcement learning, domain randomization \cite{weng2019DR, tobin2017domain,tobin_presentation} is a simple form of PCG and one way to counter overfitting in machine learning. The main idea of domain randomization is to train an agent in many  simulated training environments, where certain properties are different in each environment. The goal is to learn a single policy that can work well across all of them. In addition to trying to encourage machine learning systems to be more robust and general, another use case of domain randomization is to facilitate the transfer of policies trained in a simulator to the real world \cite{tobin2017domain,sadeghi2016cad2rl,tremblay2018training,prakash2019structured}. Training in a simulation instead of the real world has several advantages such as the training being faster, cheaper and more scalable, and having access to the ground truth. 


In a promising demonstration of this approach, Tobin et al.~\cite{tobin2017domain} trained an object detector on thousand of examples of objects with randomized colors, textures, camera positions, lighting conditions, etc. in a simulator and then showed it can detect objects in the real world without any additional training. Another example is the work by Sadeghi et al.~\cite{sadeghi2016cad2rl}, who trained a vision-based navigation policy for a  quadrotor entirely in a simulated environment with highly randomized rendering settings and then transferred this policy to the real world without further training. 

Following Weng~\cite{weng2019DR}, we can further divide domain randomization into three  subgroups: \emph{uniform domain randomization}, \emph{guided domain randomization},  and \emph{automatic domain randomization}. In uniform domain randomization, each parameter is uniformly sampled within a certain range. For example, in the work by Tobin et al.~\cite{tobin2017domain},  the size of objects, their mass, or the amount of noise added to the camera image were drawn from a uniform distribution.

\emph{Guided domain randomization.}  
In the more sophisticated guided domain randomization, the type of randomization is influenced by its effect on the training process \cite{yu2018policy,cubuk2018autoaugment,weng2019DR,zakharov2019deceptionnet}. The goal of this guided randomization is to save computational resources by focusing the training on aspects of the task that actually increase the generality of the model. 
For example, instead of randomly applying pre-defined and hard-coded data augmentation methods, the approach AutoAugment  \cite{cubuk2018autoaugment} can learn new data augmentation techniques. These augmentation techniques are optimized for based on their validation accuracy on the target dataset. Such methods can be seen as a form of adaptive content generation; in the PCG literature there are approaches to PCG that adapt to an agent-driven by e.g.\  Schmidhuber's theory of curiosity~\cite{schmidhuber1991curious,togelius2008experiment}. 

Another related approach is DeceptionNet  \cite{zakharov2019deceptionnet}, which is trained to find modifications to an image through distortion, changing the background, etc.\ that make it harder for an image recognition network to output the correct classification. Both a recognition and deception network are alternatively trained such that the deception module becomes better in confusing the recognition module, and the recognition module becomes better in dealing with the modified images created by the deception module.

\emph{Automatic Domain Randomization (ADR).} Very recently, OpenAI showed that a neural network that controls a five-fingered humanoid robot hand to manipulate a Rubik's cube, can sometimes solve this task in the real world even though it was only  trained in a simulated environment \cite{openai2019solving}. Key to this  achievement in robotic manipulation was training the robot in simulation on a large variety of different environmental variations, similar to the domain randomization approaches mentioned above. Following  related work in PCG for games \cite{justesen2018illuminating}, the ingredient to make this system work was to increase the amount  of domain randomization, as the robot gets better and better at the task. For example, while the network was initially only tasked to control a Rubik's cube of 5.7 cm, later in training it had to deal with cubes that could range from 5.47 to 6.13 cm in simulation. Because the robot had to deal with many different environments, dynamics of meta-learning did emerge in the trained neural network; this allowed the robot to adapt to different situations during test time, such as the transfer to the real world. ADR is similar to guided domain randomization but focuses more on increasing the diversity of the training environments based on task performance, instead of sampling efficiently from a  distribution of environments. 

While current domain randomization methods are showing promising results, the PCG community has  invented many sophisticated algorithms that -- we believe -- could greatly improve the generality of machine learning methods even further. As we discuss in the following sections, more recent work in PCG has focused on search-based approaches and on learning the underlying PCG representations  through machine learning techniques. 

\section{AI-driven Procedural Content Generation}

Given the successes of PCG in existing video games, as well as the perceived limitations of current PCG methods, the last decade has seen a new research field form around game content generation. The motivations include being able to generate types of game content that cannot currently be reliably generated, making game developments easier and less resource-intensive, enabling player-adaptive games that create content in response to player actions or preferences, and generating complete games from scratch. Typically, the motivations center on games and players, however, as we shall see, many of the same methods can be used for creating and varying environments for developing and testing AI.

While there has been recent work on constructive methods, more work has focused on approaches based on search and/or machine learning.

\subsection{Search-based PCG}
\label{sec:search_PCG}
In search-based PCG (Table~\ref{tbl:overview}), stochastic search/optimization algorithms are used to search for good content according to some evaluation function~\cite{togelius2011searchbased}. Often, but not always, some type of evolutionary algorithms is used, due to the versatility of these algorithms. Designing a successful search-based content generation solution hinges on designing a good representation, which enables game content to be searched for. The representation affects, among other things, which algorithms can be used in the search process; if the content can be represented as a vector of real numbers, this allows for very strong algorithms such as CMA-ES~\cite{hansen2001completely} and differential evolution~\cite{storn1997differential} to be used. If the representation is e.g.\ a graph or a permutation, this poses more constraints on the search. 
\begin{figure*}
\centering
\includegraphics[width=1.0\textwidth]{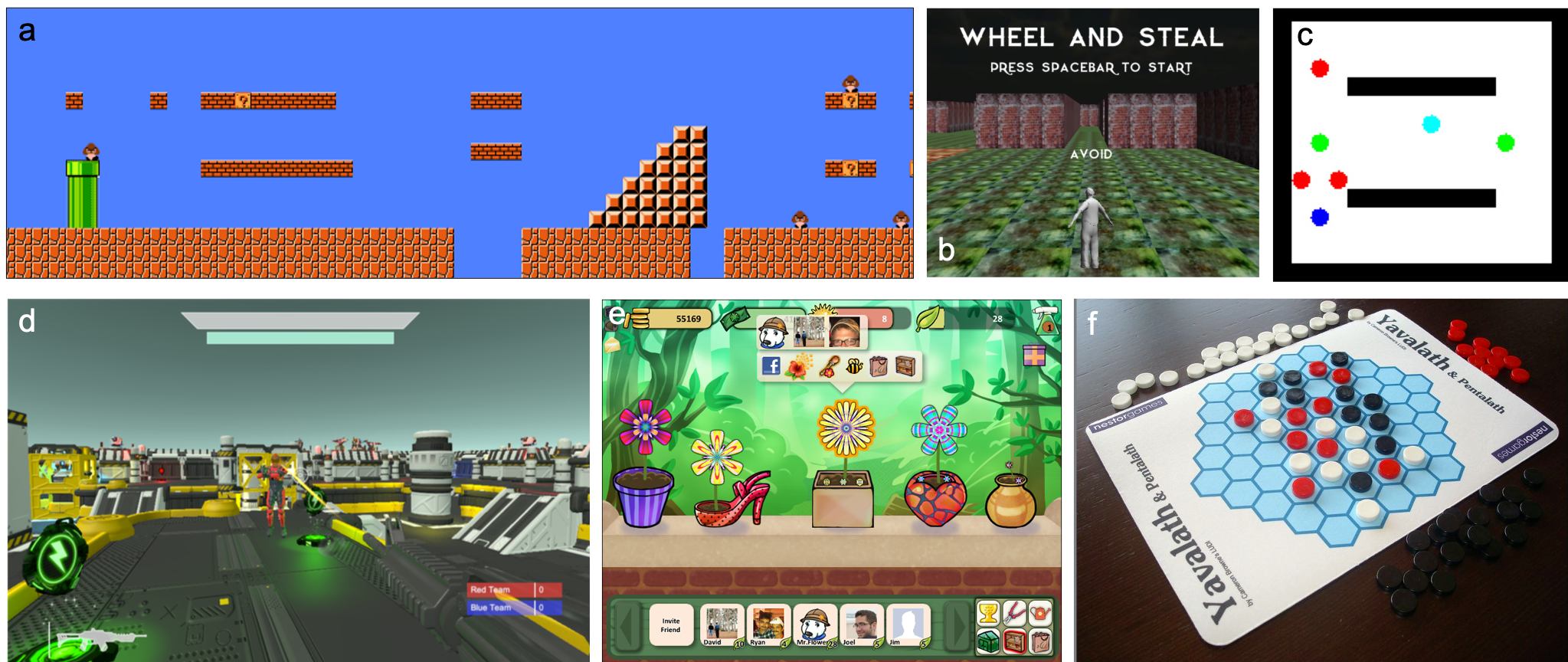}
\caption{In academia, PCG approaches have been used to produce complete and playable  (\textbf{a}) Super Mario Bros. levels \cite{summerville2016super}, (\textbf{b}) 3D \cite{cook2016angelina}, (\textbf{c}) 2D games \cite{togelius2008experiment}, and (\textbf{d}) maps and character classes for first-person shooters \cite{karavolos2019multi}. Other PCG-enabled games include Petalz (\textbf{e}), in which players can collaboratively breed an unlimited variety of different procedurally generated flowers \cite{risi12015petalz}.
The  game Yavalath (\textbf{f}) is one of the few examples of PCG-generated games that are commercially available.}
\label{fig:pcg_academia}
\end{figure*} 

An early success for search-based PCG is Browne and Maire's work on generating board games, using a game description language capable of describing rules and boards for classical board games~\cite{browne2010evolutionary}. The initial population was seeded with a dozens of such games, including \emph{Checkers}, \emph{Connect Four}, \emph{Gomoku}, and \emph{Hex}. The evaluation function was simulation-based; candidate games were evaluated through being played with a Minimax algorithm combined with a state evaluation function automatically derived for each game. The actual game evaluation is a combination of many metrics, including how often the game leads to a draw, how early in the game it is possible to predict the winner, and number of lead changes. This process, though computationally very expensive, came up with at least one game (\emph{Yavalath}), which was of sufficient quality to be sold commercially (Fig.~\ref{fig:pcg_academia}f).

While attempts at creating complete video games including game rules through search based-methods have met mixed success~\cite{togelius2008experiment,cook2011multi,cook2016angelina,nielsen2015towards}, search-based PCG has been more effective in generating specific types of game content such as levels. We have seen applications to generating maps for the real-time strategy game StarCraft~\cite{togelius2013controllable}, and levels for the platform game Super Mario Bros~\cite{dahlskog2014multi}, the first-person shooter Doom~\cite{cachia2015multi}, and the physics puzzle game Angry Birds~\cite{calle2019free}, among many similar applications. Search-based PCG has also been used for other types of game artifacts, such as particle effects for weapons~\cite{hastings2009automatic}, role-playing game classes~\cite{pantaleev2012search} and flowers~\cite{risi12015petalz} (Fig.~\ref{fig:pcg_academia}e).

The most important component in a search-based content generation pipeline is the evaluation function, which  assigns a number (or vector) to how desirable the artifact is. In many cases, this is accomplished through playing through the content in some way and assigning a value based on characteristics of the gameplay, as in the Ludi example above; other evaluation functions can be based on directly observing the artifact, or on some machine-learned estimate of e.g.\ player experience.

An emerging trend is to go beyond optimizing for a single objective, and instead trying to generate a diverse set of artifacts that perform well. The goal here is to generate, for example, not just a single level but a set of levels that vary along various dimensions, such as number of enemies of difficulty to solve for an A* algorithm~\cite{gravina2019procedural}. The Map-Elites algorithm in particular, originally introduced to create more robust robot gaits~\cite{cully2015robots}, has been adapted to create sets of game levels that vary in what skills they require from the agent or what mechanics they feature~\cite{khalifa2019intentional}.

An alternative to stochastic optimization algorithms is to use constraint satisfaction methods~\cite{smith2011tanagra}) such as answer set programming~\cite{smith2011answer}. Casting artifacts as answer sets can allow very efficient search for content that obeys specific constraints, but is hard to integrate with simulation-based evaluation methods. This paradigm is sometimes called \emph{solver-based} PCG.

\subsection{PCG via machine learning}
\label{sec:pcgml}


Machine learning methods such as Generative Adversarial Networks (GANs) \cite{goodfellow2014generative} have revolutionized the way we generate pictorial content, such as images of faces, letters, and various objects. However, when generating game content with some form of playability constraints (such as levels, maps, or quests), things become more complicated because these types of content are in some ways more like code than like images. An image of a face where the contours smudge just looks slightly off, whereas a level for Super Mario Bros with an impossibly long jump is not just somewhat defect, it's unplayable and therefore worthless. Similar functionality requirements can be found in level-like artifacts such as robot path planning problems, logic puzzles, and quests~\cite{summerville2018procedural}. Therefore we call such content, in which some algorithmic way of verifying their functionality (e.g.\ playability) exists, \emph{functional content}. 

Simply training a GAN on a large set of functional artifacts does not guarantee that the generator network learns to produce levels that fulfill these functionality requirements, nor that the discriminator learns to identify and check for those constraints. The result is often simply artifacts that look right but don't function well~\cite{torrado2019bootstrapping}. Another potential reason for the failure of machine learning-based methods to generate functional content is that methods such as GANs mostly learn local dependencies, whereas functionality in many types of content can depend on features that are far from each other, and/or counting the number of instances of a feature.

The same effect has been found with other representations, such as LSTM networks~\cite{summerville2016mystical} and Markov chains~\cite{snodgrass2016controllable}. One way of counteracting this effect is bootstrapping, where newly generated artifacts that are found to satisfy the functionality requirements are added back to the training set for continued training, thus biasing training specifically to functional artifacts~\cite{torrado2019bootstrapping}.

Machine learning models can also be combined with search to improve their efficiency. One way to do this is to use the learned model as a representation for search-based PCG. The idea here is to use machine learning to find the general space of content which is roughly defined by the examples the model is trained on, and then search within that space. Using GANs, this could be done by searching the \emph{latent space}; when training a GAN, a latent vector is used as input to the generator network. The latent space is defined by that input. \emph{Latent Variable Evolution} refers to using evolutionary algorithms to search the latent space for artifacts that optimize some kind of objective function~\cite{bontrager2018deepmasterprints}. For example, latent variable evolution was used to generate new levels for Super Mario Bros, by first training a GAN on one-screen segments of most levels in the original game. The latent space was then searched for vectors that would maximize objectives such as that the segment should contain many jumps, or should not be winnable without jumping, or should be unwinnable~\cite{volz2018evolving}. 

Functionality evaluation can be integrated into adversarial learning processes in other ways.
Generative Playing Networks consist of a generator network, that generates levels, and a reinforcement learning agent that learns to play them~\cite{bontrager2020fully}. While the objective for playing agent is simply to perform as well as possible on the level, the objective for the playing agent is to provide an appropriate level of challenge for the agent.

Another way of using machine learning for PCG is to use reinforcement learning. The conceptual shift here is to see PCG as a sequential process, where each action modifies a content artifact in some way. The goal of the training process then becomes to find a policy that for any content state selects the next action so that it leads to maximum expected final content quality. For this training process to be useful, we will need the trained policy to be a content generator capable of producing diverse content, rather than simply producing the same artifact every time it is run. A recent paper articulates a framework for PCG via reinforcement learning and proposes methods for ensuring that the policy has sufficiently diverse results in the context of generating two-dimensional levels~\cite{khalifa2020pcgrl}. Two important lessons learned is to always start from a randomized initial state (which need not be a functional level) and to use short episodes, to prevent the policy from always converging on the same final level. (It is interesting to note that the issues with learning general policies in RL recur in trying to learn policies that create content that can help generalize RL policies.)

Compared to PCG based on supervised or self-supervised learning, PCG based on reinforcement learning has the clear advantage of not requiring prior content to train on, but the drawback of requiring a reward function judging the quality of content. This is very similar in nature to the evaluation function in search-based PCG. Compared to search-based PCG, PCG via reinforcement learning moves the time and computation expense from inference to training stage; whereas search-based PCG uses extensive computation in generating content, PCG via reinforcement learning uses extensive computation to train a model which can then be used cheaply to produce additional content.

\section{Procedurally generated learning environments}

\begin{figure*}
\centering
\includegraphics[width=1.0\textwidth]{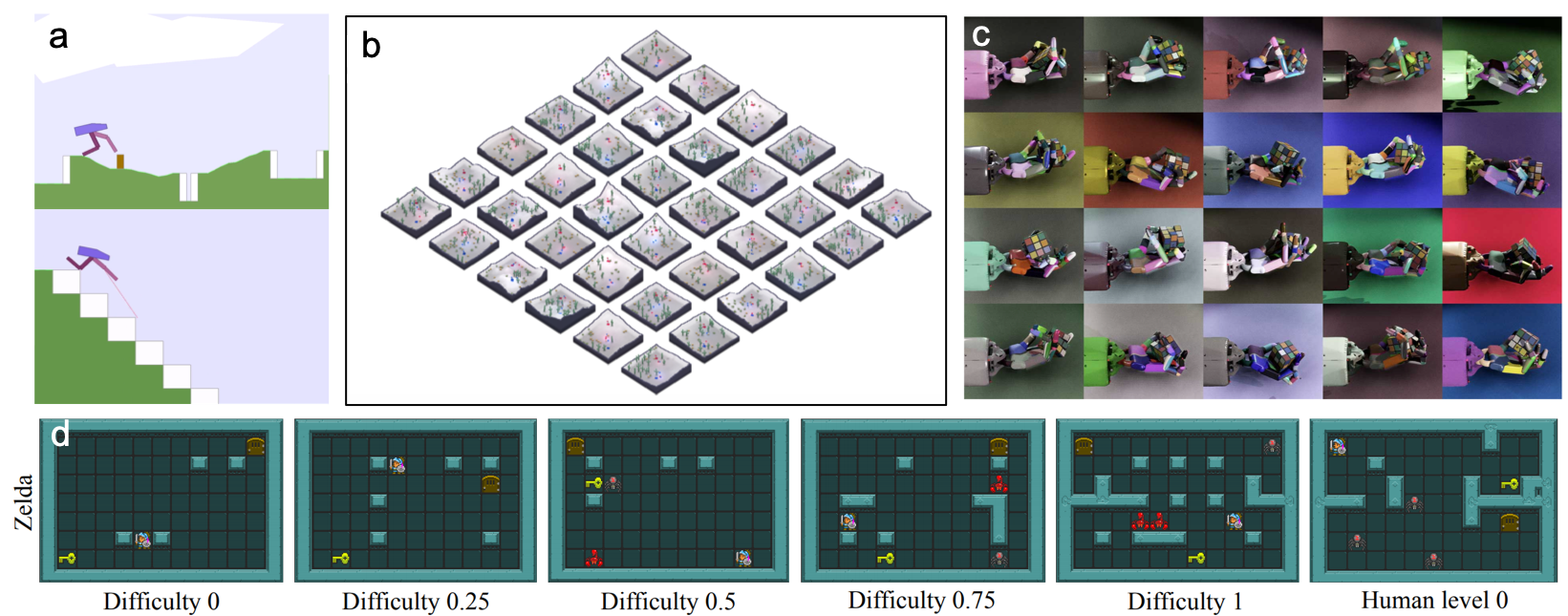}
\caption{Examples of learning environments created by PCG-based approaches. The POET algorithm (\textbf{a}) learns to create increasingly complex environments for a 2D bipedal walker together with their neural network controllers   \cite{wang2019paired}. (\textbf{b}) Procedurally generated maps were one of the key ingredients to allow agents to master the Quake III Capture the Flag domain. Increasing task complexity depending on the performance of the agent has shown to lead to more general solutions for (\textbf{c}) controlling a robot hand to manipulate a Rubik's cube in simulation and in the real world \cite{openai2019solving}, and  (\textbf{d}) video game playing \cite{justesen2018illuminating}.}
\label{fig:learning_env}
\end{figure*}

An exciting opportunity of PCG algorithms is to create the actual learning environments that scaffold the learning of artificial agents (Fig.~\ref{fig:learning_env}). Similarly to how current machine learning methods are moving towards automating more and more facets of training (e.g.\ meta-learning the learning algorithms themselves, learning network architectures instead of hand-designing them), the automated generation of these progressive curricula that can guide learning offers unique benefits.

One of the first examples of this idea is \emph{Minimal Criterion Coevolution} (MCC) \cite{brant2017minimal}. In MCC both the agent and the environment co-evolve to solve increasingly more difficult mazes. Recent work building on these ideas is POET \cite{wang2019paired}, which deals with the more challenging OpenAI gym bipedal walker domain. POET is a good example of an approach in which solutions to a particular obstacle-course can function as stepping stones for solving another one. In fact, for the most difficult environments (shown on the right in Fig.~\ref{fig:learning_env}a) it was not possible to directly train a solution; the stepping stones found in other environments were necessary to solve the most ambitious course. 

Importantly, procedurally generated training environments can also increase the generality of reinforcement learning agents. Zhang et al.~\cite{zhang2018study} showed that training on thousands of levels in a simple video game can allow agents to generalize to levels not seen before. Some domains in the OpenAI gym training environments include procedurally generated content, requiring the agents to learn more general strategies.  For example, in the CarRacing-v0 environment \cite{oleg2016}, agents are presented with a new procedurally generated car racing track  every episode and the final reward is the average reward over multiple rollouts. These procedurally generated environments required more sophisticated neural architectures to be solvable \cite{ha2018recurrent}, highlighting their usefulness in testing the ability of the reinforcement learning agents. A similar approach for encouraging the discovery of general policies worked well for evolving stable policies for a 2D bipedal walker domain \cite{ha2017stable}. In addition to helping in supervised learning settings (see Section~\ref{sec:augment}), forms of data augmentation can also help RL agents to become more robust. Randomized environments are also present in the Arena multi-agent testbed~\cite{lukasiewicz2020arena}. In the work by Cobbe et al.~\cite{cobbe2018quantifying} agents are trained in environments in which random rectangular regions of the environment are cut out and replaced by rectangles filled with random colors, which helps these agents to generalize better.

Jaderberg et al.~\cite{jaderberg2019human} relied on a PCG-based approach to allow RL agents to master Quake III Arena Capture the Flag (Fig.~\ref{fig:learning_env}b). In their work, agents were trained on a mixture of procedurally generated indoor and outdoor maps (with varying walls and flag locations), which allowed the agents to learn policies that are robust to variations in the maps or the number of players. This work also demonstrated another advantage of procedurally generated maps: because each map is different, agents learned to learn how to keep track of particular map locations (e.g.\ the entrance to the two bases) through their external memory system.

While the aforementioned work \cite{zhang2018study,jaderberg2019human} showed that training on a larger variety of environments can lead to more general agents, it did require a large number of training levels. In work from the PCG research community, Justesen et al.~\cite{justesen2018illuminating}  introduced a \emph{Progressive PCG} approach (PPCG), which showed that performance of training agents can be increased while using less data if the difficulty of the level is changed in response to the performance of the agents (Fig.~\ref{fig:learning_env}d). A similar approach was then later adopted by OpenAI to train their humanoid robot hand (Fig.~\ref{fig:learning_env}c) in increasingly more challenging environments \cite{openai2019solving}. 

In fact, there is some initial evidence that very varied training environments can also foster the emergence of meta-learning in recurrent neural networks, that allow adaption to situations not seen during training \cite{openai2019solving}.     
While approaches such as OpenAI's Rubik's cube solving robot hand  hint at the potential of this approach, creating an encoding that can produce an even larger variety of different and effective training environments could have a significant impact on the generality of the  agents and robots we are able to train.  

We also summarise the similarities and differences between POET, MCC, and PPCG in Table~\ref{tbl:overview}. While all three approaches use hand-designed representations, PPCG does not evolve the levels but instead uses a ruled-based generator.

\section{Looking forward: Opportunities and Challenges}

With more advanced ML techniques, PCG approaches are becoming better and better at generating content while PCG methods are now starting to allow more general machine learning systems to be trained. We believe the idea of automatically and procedurally generating learning environments with the right complexity that scaffold the learning of autonomous agents is an exciting research direction that can help overcome some of the constraints that impede generalization and open-ended learning in current AI. This research direction is similar to what has been proposed in PCG research before, and also to the idea of AI-generating algorithms (AI-GA)~\cite{clune2019ai}. We identify five main open challenges  that we believe are essential in pushing the field of PCG forward and to realizing its promise to create more adaptive and lifelong learning ML agents. 

\subsection{Learning from limited data}
When generating images with machine learning, it is common practice to train the model on thousands, maybe even millions, of images~\cite{lecun2015deep}. However, such amounts of high-quality data are rarely available when developing a game, or even in a finished game. For example, the original \emph{Super Mario Bros} game has 32 levels, resulting in a few hundred screens' worth of content. For some games, a large amount of user-generated content is available online, but this content can be of very variable quality. And when creating, for example, a robot learning benchmark from scratch, creating scenarios to train a content model on can be a significant time investment. Bootstrapping in PCG \cite{torrado2019bootstrapping}
 (see Section~\ref{sec:pcgml}) can help overcome this problem of content shortage, and various data augmentation could also help but learning to generate new content from limited data is still a significant challenge. More research is needed on how to learn from little data, and on how to learn generative models based on many different types of data. For example, training a model on lots of available benchmark rules to learn generic patterns, it should be possible to generate environments for a new benchmark.

\subsection{Generating Complete Games}
\label{generatingcompletegames}
While PCG techniques have shown impressive results for particular types of content in particular game genres, there has been much less progress on the harder problem of generating complete games. Browne and Maire's work from 2010 (discussed above~\cite{browne2010evolutionary}), which resulted in a well-reviewed board game which is sold in stores, remains the gold standard. Generating complete video games~\cite{togelius2008experiment,nielsen2015towards,smith2011answer,nelson2007towards,cook2011multi,cook2016angelina} (Fig.~\ref{fig:pcg_academia}) or card games~\cite{font2013card} seems to be much harder challenges, with the results often being unplayable or uninteresting. Methods that have been tried include constraint satisfaction through answer set programming as well as evolutionary search. This is partly because of these games being very complex, and partly because it is very hard to find good evaluation metrics for complete games. Yet, generating complete challenges, including rules, topology, visuals etc, seems a crucial part of a process where we gradually scale up challenges for agents that are capable of completing not just one challenge, but multiple ones. 

PCG via machine learning could be a potentially promising approach to tackle this challenge. For example, Fan et al.~\cite{fan2019generating} very recently showed that a neural network can learn from crowd-sourced elements such as descriptions of locations and characters to create multiplayer text adventure games. This idea of leveraging and integrating real world data to create games (also known as \emph{Data Games}), was first proposed by Gustafsson et al.~\cite{gustafsson2013data} and later extended to procedurally generate simple adventures games using open data from Wikipedia \cite{barros2016playing}. Another example of how to leverage advances in machine learning for PCG is the recent AI Dungeon 2 text adventure game \cite{aidungeon2020}. In this game, players can type in any command and the system can respond to it reasonably well, creating the first never-ending text adventure. The system is built on OpenAI's GPT-2 language model \cite{radford2019language}, which was further fine-tuned on a number of text adventure stories. This work also highlights that machine learning techniques combined with PCG might lead to completely new types of games that would not have been possible without advanced AI methods.  

\subsection{Lifelong generation for lifelong learning}

The problem of Lifelong Learning is that of continuously adapting and improving skills over a long lifetime of an agent, comprising many individual episodes, though not necessarily divided into episodes as currently thought of~\cite{thrun1995lifelong,parisi2019continual}. This would require for an agent to build on previously learned skills as it faces increasingly harder or more complex, or just more varied, challenges. Lifelong learning is a problem, or maybe rather a setting, whose popularity has seemingly waxed and waned (under different names) as subsequent generations of researchers have discovered this challenge and then understood how hard it is. Within the artificial life community, the challenge of simulating  \emph{open-ended evolution} is closely related to that of lifelong learning. The idea behind open-ended evolution is to try to computationally  replicate the process that allows nature to endlessly produce a diversity of interesting and complex artifacts. Environments such as Tierra~\cite{ray1993evolutionary} and  Avida~\cite{adami1994evolutionary} were early attempts at realizing that possibility. 

The procedural generation of environments and challenges is a great opportunity for lifelong learning, and might even be a precondition for lifelong learning to be practically possible. It is possible that earlier attempts at realizing lifelong learning have had limited success partly because the environments lacked sufficient challenges of the right complexity. The POET system shows one way of co-creating environments with agents~\cite{wang2019paired}. However, there is a great outstanding research challenge in devising mechanisms for gradually growing or complexifying environments (see next Section) so as to generate the right problems at the right time for agents to continually learn.

\subsection{New PCG-based RL benchmarks}

A variety of benchmarks have been proposed to test the generalization abilities of RL algorithms. Justesen et al.~\cite{justesen2018illuminating} used procedurally generated levels in the  General Video Game AI (GVG-AI) framework \cite{perez2018general}, to study overfitting of RL algorithms to different level distribution.  In a similar vein to the work by Justesen et al.~\cite{justesen2018illuminating}, levels in the CoinRun platform game are procedurally generated to quantify the ability of RL algorithms to generalize to never-before-seen levels \cite{cobbe2018quantifying,cobbe2019leveraging}. Another procedurally generated environment is the Unity game engine-based Obstacle Tower environment \cite{juliani2019obstacle}, which requires increasingly complex skills such as  locomotion, planning, and puzzle-solving. Others have recently combined the Unity environment with GVGAI, creating UnityVGDL \cite{johansen2019video}, which allows ML agents in Unity to be tested on a large selection of games.  

Other setups that do not use  PCG include the work by  Nichol et al.~\cite{nichol2018gotta}, in which  \emph{Sonic the Hedgehog$^{TM}$} levels were separated into a training and test set to investigate how well RL algorithms generalize. In the Psychlab environment \cite{leibo2018psychlab}, agents are tested on known tasks from cognitive psychology, such as visual search or object tracking, making the  results from simulated agents directly comparable to human results.  


We propose the creation of PCG-based benchmarks in which the agent's environment and reward is non-stationary and becomes more and more complex over time. A starting point could be PCG approaches that are able to evolve the actual rules of a game (see section \ref{generatingcompletegames}). New rules could be introduced based on agents' performance and estimates of their learning capacity. 

Adaptation within trials is as important as adaptation between trials: a generator could generate increasingly difficult games, which are different enough in each trial that a policy that would not adapt within a trial would fail. The Animal-AI Environment \cite{beyret2019animal}, in which agents have to adapt to unforeseen challenges based on classical tests from  animal cognition studies, shares similar ideas with the benchmarks we are proposing here but does not focus on  procedurally generated environments and tasks.


\subsection{From simulation to the real world}
Procedurally generated environments have shown their potential in training robot policies that can cross the reality gap. Promising work  includes approaches that try to learn the optimal  parameters of a simulator, so that policies trained in that simulator work well with real data \cite{ruiz2018learning, kar2019meta}. However, current approaches are still limited to lab settings, and we are far from being able to train robots that can deal with the messiness and diversity of tasks and environments encountered in the real world. 

An intriguing opportunity  is to train policies in much more diverse simulated environments than have been explored so far, with the hope that they will be able to cope better with a wider range of tasks when transferred to real physical environments. Both the Unity Simulation environment and Facebook's AI Habitat are taking a step in this direction. With Unity Simulation, Unity is aiming for simulation environments to work at scale, allowing developers to built digital twins of factories, warehouses or driving environments.  Facebook's AI Habitat is designed to train embodied agents and robots in photo-realistic 3D environments to ultimately allow them to work in the real world. 

In addition to developing more sophisticated machine learning models, one important research challenge in crossing the reality gap is the \emph{content gap} \cite{kar2019meta}. Because the synthetic content that the agents are trained on typically only represents a limited set of scenarios that might be encountered in the real world, the agents will likely fail if they encounter situations that are too different from what they have seen before. 

How to create PCG approaches that can limit this content gap and create large and diverse training environments, which prepare agents well for the real world tasks to come, is an important open research direction.  

\section*{Acknowledgements}
We would like to thank all the members of modl.ai, especially Niels Justesen, for comments on earlier drafts of this manuscript. 
We would also like to thank Andrzej Wojcicki, Rodrigo Canaan, Nataniel Ruiz, and the anonymous reviewers for additional comments and suggestions. Both authors (SR and JT) contributed equally to the conceptualization and writing of the paper.

\bibliographystyle{unsrt}
\bibliography{references}

\end{document}